\documentclass[12pt]{amsart}
\usepackage{amsmath}
\usepackage{amssymb,amsfonts}
\usepackage[dvips]{graphicx}
\usepackage{graphicx}
\usepackage{color}
\usepackage{epsfig,amscd}
\usepackage{latexsym}
\usepackage{natbib}
\theoremstyle{plain}
\newtheorem{thm}{Theorem} 

\newtheorem{lemma}[thm]{Lemma}
\newtheorem{comment}{Comment}
\theoremstyle{definition}

\newtheorem{corollary}{Corollary}

\newcommand\e{\epsilon}

\begin{document}

\title{Filtering additive measurement noise\\with maximum entropy in the mean}

\author{Henryk Gzyl, Enrique ter Horst}
\address{Facultad de Ciencias \\
Universidad Central de Venezuela \\
Instituto de Estudios Superiores de Administraci\'on IESA\\
Caracas, DF, Venezuela.}
\email{hgzyl@reacciun.ve, enrique.terhorst@iesa.edu.ve}

\begin{abstract} The purpose of this note is to show how the method of maximum entropy in the mean (MEM) may be used to improve parametric estimation when the measurements are corrupted by large level of noise. The method is developed in the context on a concrete example: that of estimation of the parameter in an exponential distribution. We compare the performance of our method with the bayesian and maximum likelihood approaches.

\end{abstract}

\maketitle

\baselineskip=24pt

\section{Introduction}
Suppose that you want to measure the half-life of a decaying nucleus or the life-time of some elementary particle, or some other random variable modeled by an exponential distribution describing, say a decay time or the life time of a process. Assume as well that the noise in the measurement process can be modeled by a centered gaussian random variable whose variance may be of the same order of magnitude as that of the decay rate to be measured. To make things worse, assume that you can only collect very few measurements.

That is if $x_i$ denotes the realized value of the variable, one can only measure $y_i = x_i + e_i$, for $i = 1,2,...,n$, where $n$ is a small mumbler, say $2$ or $3,$ and $\e_1$ denotes the additive measurement noise. In other words, assume that you know that the sample comes from a specific parametric distribution but is contaminated by additive noise. What to do?  One possible approach is to apply small sample statistical estimation procedures. But these are designed for problems where the variability is due only to the random nature of the quantity measured,and there is no other noise in the measurement

Still another possibility, the one we that to explore here, is to apply a maxentropic filtering method,  to estimate both the unknown variable and the noise level. For this we recast the problem as a typical inverse problem consisting of solving for $\mathbf{x}$ in
\begin{equation}\label{inverse_prob}
\bf{y = Ax + e};\;\;\;x \in K\rm
\end{equation}

\noindent where $\mathbf{K}$ is a convex set in $\mathbb{R}^d$, $\,\mathbf{y} \in \mathbb{R}^k$ and for some $d$ and $k$, and $\bf{A} \rm$ is an $k\times d$-matrix which depends on how we rephrase the our problem. We could, for example, consider the following problem: Find $\hat{x} \in [0,\infty)$ such that

\begin{equation}\label{problem}
\hat{y} = \hat{x} + \hat{e}
\end{equation}

 In our case $\bf{K}\rm = [0,\infty)$, and we set $\hat{y} = \frac{1}{n}\Sigma_j\,y_j.$ Or we could consider a collection of $n$ such problems, one for every measurement, and then proceed to carry on the estimation. Once we have solved the generic problem (\ref{inverse_prob}), the variations on the theme are easy to write down. What is important to keep in mind here, is that the output of the method is a filtered estimator $\hat{x}^*$ of $\hat{x},$ which itself is an estimator of the unknown parameter. The novelty then is to filter out the noise in (\ref{problem}).

The method of maximum entropy in the mean is rather well suited for solving problems like (1). See Navaza (1986) for an early development and Dacunha-Castele and Camboa (1990) for full mathematical treatment . Below we shall briefly review what the method is about and then apply it to  obtain an estimator $\hat{x}$ from (\ref{problem}). In section 3  obtain the maxentropic estimator and in section 4 we examine some of its properties, in particular we examine what the results would be if either the noise level were small or the number of measurements were large. We devote section 4 to some simulations in which the method is compared with a bayesian and a maximum likelihood approaches.

\section{The basics of MEM}

	MEM is a technique for transforming a possibly ill-posed, linear problem with convex constraints into a simpler (usually unconstrained) but non-linear minimization problem. The number of variables in the auxiliary problem being equal to the number of equations in the original problem, $k$ in the case of example 1. To carry out the transformation one thinks of the $\mathbf{x}$ there as the expected value of a random variable $\mathbf{X}$ with respect to some measure $\mathbb{P}$ to be determined. The basic datum is a sample space $(\Omega_s, \mathcal{F}_s)$ on which $\mathbf{X}$ is to be defined. In our setup the natural choice is to take $\Omega_s = \mathbf{K}$, $\mathcal{F}_s = \mathcal{B}(\mathbf{K})$, the Borel subsets of $\mathbf{K}$, and $\mathbf{X} = \mathbf{id}_{\mathbf{K}}$ the identity map. Similarly, we think of $\mathbf{e}$ as the expected value of a random variable $\mathbf{V}$ taking values in $\mathbb{R}^k$. The natural choice of sample space here is $\Omega_n = \mathbb{R}^k$ and $\mathcal{F}_n = \mathcal{B}(\mathbb{R}^k)$ the Borel subsets.

To continue we need to select to prior measures $dQ_s(\xi)$ and $dQ_n(v)$ on $(\Omega_s,\mathcal{F}_s)$ and $(\Omega_n,\mathcal{F}_n)$. The only restriction that we impose on them is that the closure of the convex hull of both $supp(Q_s)$ (resp. of $supp(Q_n)$) is $\mathbf{K}$ (resp. $\mathbb{R}^k$). These prior measures embody knowledge that we may have about $\mathbf{x}$ and $\mathbf{e}$ but are not priors in the Bayesian sense. Actually, the model for the noise component describes the characteristics of the measurement device or process, and it is a datum. The two pieces are put together setting $\Omega = \Omega_s \times \Omega_n$; $\mathcal{F} = \mathcal{F}_s\otimes \mathcal{F}_n$, and $dQ(\xi,v) = dQ_s(\xi)dQ_n(v)$. And to get going we define the class
\begin{equation}
\mathbb{P} = \{P\,|\,P << Q;\; AE_P[\mathbf{X}] + E_P[\mathbf{V}] = \mathbf{y}\}.
\end{equation}

Note that for any $P \in \mathbb{P}$ having a strictly positive density $\rho = \frac{dP}{dQ}$, then $E_P[\mathbf{X}] \in \rm int(\mathbf{K})$. For this standard result in analysis check in Rudin's (1973) book. The procedure to explicitly produce such $P$'s is known as the maximum entropy method. The first step of which is to assume that  $\mathbb{P} \not= \emptyset$,  which amounts to say that our inverse problem (1) has a solution and define
$$
 S_Q: \mathbb{P} \rightarrow [-\infty, \infty)\\
$$
by the rule
\begin{equation}
S_Q(P) = -\int_\Omega \ln(\frac{dP}{dQ})dP
\end{equation}
\noindent whenever the function $\ln(\frac{dP}{dQ})$ is $P$-integrable and $S_Q(P) = -\infty\;$ otherwise. This entropy functional is concave on the convex set $\mathbb{P}$. To guess the form of the density of the measure $P^*$  that maximizes $S_Q$ is to consider the class of exponential measures on $\Omega$ defined by
\begin{equation}
dP_\lambda =\frac{e^{-<\lambda,\mathbf{A}\xi>-<\lambda,v>}}{Z(\lambda)}dQ
 \end{equation}
\noindent where the normalization factor is
$$
Z(\lambda) = E_Q[e^{-<\lambda,\mathbf{A}\xi>-<\lambda,v>}].
$$
Here $\lambda \in \mathbb{R}^k$. If we define the dual entropy function
$$
\Sigma(\lambda) : \mathcal{D}(Q) \rightarrow (-\infty,\infty]
$$
\noindent by the rule
\begin{equation}
\Sigma(\lambda) = \ln Z(\lambda) + <\lambda,\mathbf{y}>
\end{equation}
\noindent or $\Sigma(\lambda) = \infty$ whenever $\lambda \notin \mathcal{D}(Q)\equiv \{\mu \in \mathbb{R}^k\;|\; Z(\mu) < \infty\}$.

It is easy to prove that, $\Sigma(\lambda) \geq S_Q(P)$ for any $\lambda \in \mathcal{D}(Q)$, and any $P \in \mathbb{P}$. Thus if we were able to find a $\lambda ^* \in \mathcal{D}(Q)$ such that $P_{\lambda^*} \in \mathbb{P}$, we are done. To find such a $\lambda^*$ it suffices to minimize (the convex function) $\Sigma(\lambda)$ over (the convex set) $\mathcal{D}(Q)$. We leave for the reader to verify that if the minimum is reached in the interior of $\mathcal{D}(Q)$, then $P_{\lambda^*} \in \mathbb{P}$. We direct the reader to Borwein and Lewis (2000) for all about this, and much more.

\section{Entropic Estimators}

Let us now turn our attention to equation (2). Since our estimator is a sample mean of an exponential (of unknown parameter) it is natural to assume for the method described in section 2, to assume that the prior $Q_s$ for $\mathbf{X}$ is a $\Gamma(n,\alpha /n)$, where $\alpha > 0$ is our best (or prior) guess of the unknown parameter. Below we propose a criterion for the best choice of $\alpha.$ Similarly, we shall chose $Q_n$ to be the distribution of a $N(0,\delta /n)$ random variable as prior for the noise component.

Things are rather easy under these assumptions. To begin with, note that
$$
Z(\lambda) = \frac{e^{\frac{\lambda ^2 \delta ^2}{2n}}}{(\frac{\lambda}{n\alpha} + 1)^n}
$$
and the typical member $dP_\lambda(\xi,v)$ of the exponential family is now

\begin{equation}
 dP_\lambda(\xi,v) = (\lambda +n\alpha)^n \frac{\xi^{n-1}}{\Gamma(n)}e^{-(\lambda +n\alpha)\xi}\,\frac{e^{-(v+\frac{\delta^2 \lambda}{n})^2\frac{n}{2\delta^2}}}{(2\pi \delta^2 /n)^{1/2}}d\xi dv .
\end{equation}

	It is also easy to verify that the dual entropy function $\Sigma(\lambda)$ is given by
$$
\Sigma(\lambda)= \frac{\lambda^2 \delta^2}{2n} - n\ln (\frac{\lambda}{n\alpha}+1) + \lambda \hat{y}
$$
the whose minimum value is reached at $\lambda^*$ satisfying
\begin{equation}\label{minimizer}
\frac{\lambda^*\delta^2}{n}-\frac{1/\alpha}{\frac{\lambda^*}{n\alpha}+1} + \hat{y} = 0
\end{equation}
\noindent and, discarding one of the solutions (because it leads to a negative estimator of a positive quantity), we are left with

$$\frac{\lambda^*}{n \alpha} = \frac{1}{2}(-(1 +  \frac{\hat{y}}{\alpha\delta^2}) + ((1 -  \frac{\hat{y}}{\alpha\delta^2})^2 + \frac{4}{\alpha^2\delta^2})^{1/2})$$

\noindent from which we obtain that
\begin{equation}
\frac{\lambda^*}{n \alpha}+1 = \frac{1}{2}((1 - \frac{\hat{y}}{\alpha\delta^2}) + ((1 - \frac{\hat{y}}{\alpha\delta^2})^2 + \frac{4}{\alpha^2\delta^2})^{1/2})
\end{equation}
\noindent as well as
\begin{equation}\label{mem}
\begin{array}{rcl}
\hat{x}^* &=& E_{P(\lambda^*)}[\mathbf{X}] = \frac{n}{(\lambda^* +n\alpha)} = [\frac{\alpha}{2}\Big((1-\frac{\hat{y}}{\alpha\delta})+\sqrt{(1-\frac{\hat{y}}{\alpha\delta})^2 +\frac{4}{\alpha^2\delta^2}}\big)^{1/2}]^{-1}\\
\hat{e}^* &=& E_{P(\lambda^*)}[\mathbf{V}] = -\frac{\delta ^2 \lambda^*}{n}.
\end{array}
\end{equation}
\begin{comment}
Clearly, from (\ref{minimizer}) it follows that $\hat{y} = \hat{x}^* + \hat{e}^*$. Thus it makes sense to think of $\hat{x}^*$ as the estimator with the noise filtered out, and to think of $\hat{e}^*$ as the residual noise.
\end{comment}

\section{Properties of $\hat{x}^*$}
Let us now spell out some of the notation underlying the probabilistic model behind (1). We shall assume that the $x_i$ and the $e_i$ in the first section are values of random variables $X^i$ and $\epsilon^i$ defined on a sample space $(\mathbb{W}, \mathcal{W})$. For each $\theta > 0$, we assume to be given a probability law $P(\theta)$ on $(\mathbb{W}, \mathcal{W})$, with respect to which the sequences $\{X^k \,|\,k=1,2,...\}$ and $\{\epsilon^k \,|\,k=1,2,...\}$ are both i.i.d. and independent of each other, and that with respect to $P(\theta)$, $X^k \sim \, \exp(\theta)$ and $\epsilon^k \sim \,N(0,\delta^2)$. That is we consider the underlying model for the noise as our prior model for it. Minimal consistency is all right. Form the above, the following basic results are easy to obtain.

From (9) and (10) it is clear that

\begin{lemma} If we take $\alpha = 1/\hat{y}$, then $\lambda^* = 0$ and $\hat{x}^* = \hat{y}$ and $\hat{e}^* = 0$.
\end{lemma}
\begin{comment}Actually it is easy to verify that the solution to $\hat{x}^*(\alpha) = 1/\alpha$ is $\alpha = 1/\hat{y}.$
\end{comment}
To examine the case in which large data sets were available, let us add a superscript $n$ and write $\hat{y}{(n)}$ to emphasize the size of the sample. If $\hat{x}^{(n)}$ denotes the arithmetic mean of an i.i.d. sequence of random variables having $\exp(\theta)$ as common law, it will follow form the LLN that

\begin{lemma} As $n \rightarrow \infty$ then
\begin{equation}\label{LLN}
(\hat{x}^{(n)})^* \rightarrow \tilde{x}(\alpha) \equiv [\frac{\alpha}{2}\Big((1 - \frac{\theta }{\alpha\delta^2}) + ((1 - \frac{\theta }{\alpha\delta^2})^2 + \frac{4}{\alpha^2\delta^2})^{1/2}\Big)]^{(-1)}.
\end{equation}
\end{lemma}

\begin{proof}Start from (10), invoke the LLN to conclude that $\hat{y}(n)$ tends to $\theta$ and obtain (\ref{LLN}).
\end{proof}
\begin{corollary} The true parameter is the solution of $\tilde{x}(\alpha)- 1/\alpha =0$.
\end{corollary}
\begin{proof} Just look at the right hand side of (\ref{LLN}) to conclude that $\tilde{x}(1/\theta) = \theta$.
\end{proof}
\begin{comment} What this asserts is that when the number of measurements is large, to find the right value of the parameter it suffices to solve $\tilde{x}(\alpha)- 1/\alpha =0.$
\end{comment}
And when the noise level goes to zero, we have

\begin{lemma}With the notations introduced above, $\hat{x}^* \rightarrow \hat{y}$ as $\delta \rightarrow 0.$
\end{lemma}
\begin{proof} When $\delta \rightarrow 0$, the $dQ_n(v) \rightarrow \epsilon_0(dv)$ the Dirac point mass at $0$. In this case, we just set $\delta = 0$ in (\ref{minimizer}) and the conclusion follows.
\end{proof}
When we choose $\alpha = 1/\hat{y}$, the estimator $\hat{x}^*$ happens to be unbiased.
\begin{lemma} Let $\theta$ denote the true but unknown parameter of the exponential, and  $P_\theta(dy)$ have density
$$f_\theta(y) = \int_{-\infty}^y \theta^n (y-s)^{n-1}\frac{e^{-\theta(y-s)}e^{-s^2/2\delta^2}}{\Gamma(n)\sqrt{2\pi}\delta}ds$$
\noindent for $y > 0$ and $0$ otherwise. With the notations introduced above,  we have $E_{P(\theta)}[(\hat{x}^{(n)})^*] = 1/\theta$ whenever the prior $\alpha$ for the maxent is the sample mean $\hat{y}$.
\end{lemma}
\begin{proof} It drops out easily from Lemma 1, from (2) and the fact that the joint density $f_\theta$ of $\hat{y}$ is a convolution.
\end{proof}
But the right choice of the parameter $\alpha$ is still a pending issue. To settle it we consider once more the identity $|\hat{y} - \hat{x}^*| = |\hat{e}^*|.$ In our particular case we shall see that $\alpha = 0$ minimizes the right hand side of the previous identity. Thus, we propose to choose $'alpha$ to minimize the residual or reconstruction error.
\begin{lemma} With the same notations as above, $\hat{e}^*$ happens to be a monotone function of $\alpha$ and
$\hat{e}^*(\alpha = 0) = \frac{1}{2}\big(\hat{y}-\sqrt{\hat{y}^2 + 4\delta^2}\big)$ and $\hat{e}^*(\alpha \rightarrow \infty) = \hat{y}.$ In the first case $\hat{x}^*(\alpha = 0) = \frac{1}{2}\big(\hat{y} + \sqrt{\hat{y}^2 + 4\delta^2}\big),$ whereas in the second $\hat{x}^*(\alpha \rightarrow \infty) = 0.$
\end{lemma}
\begin{proof} Recall from the first lemma that when $\alpha\hat{y} = 1$, then $\hat{e}^* = 0$. A simple algebraic manipulation shows that when $\alpha\hat{y} > 1$ then $\hat{e}^* > 0,$ and that when $\alpha\hat{y} < 1$ then $\hat{e}^* < 0.$. To compute the limit of $\hat{e}^*$ as $\alpha \rightarrow \infty$, note that for large $\alpha$ we can neglect the term $4/\delta^2$ under the square root sign, and then the result drops out. It is also easy to check the positivity of the derivative of $\hat{e}^*$ with respect to $\alpha.$
Also clearly $|\hat{e}^*(0)| < |\hat{e}^*(\infty)|.$
\end{proof}
To sum up, with the choice $\alpha = 0$, the entropic estimator and residual error are
\begin{equation}\label{bestmem}
 \hat{x}^*(0) = \frac{1}{2}\big(\hat{y} + \sqrt{\hat{y}^2 + 4\delta^2}\big),\;\;\;\;\hat{e}^*(0) = \frac{1}{2}\big(\hat{y}-\sqrt{\hat{y}^2 + 4\delta^2}\big).
\end{equation}

\section{Simulation and comparison with the Bayesian and Maximum Likelihood approaches}

In this section we compare the proposed maximimum entropy in the mean procedure with the bayesian and maximum likelihood estimation procedures. We do that simulating data and carrying out the three procedures and plotting the histograms of the corresponding histograms. First, we generate histograms that describe the statistical nature of $\hat{x}^*$ as a function of the parameter $\alpha$. For that we generate a data set of 1000 samples, and for each of them we obtain $\hat{x}^*$ from (\ref{bestmem}).  Also, for each data point we apply both a Bayesian estimation method and a maximum likelihood method, and plot the resulting histograms.
\subsection{The maxentropic estimator}

The simulated data process  goes as follows. For $n=3$ the data points $y_1,y_2,y_3$ are obtained in the following way:

\begin{itemize}
\item Simulate a value for $x_i$ from an exponential distribution with parameter $\theta (= 1)$.
\item Simulate a value for $e_i$ from a normal distribution $N(0,\delta = 0.5)$
\item Sum $x_i$ with $e_i$ to get $y_i$, if $y_i<0$ repeat first two steps until $y_i>0$
\item Do this for $i=1,2,3$.
\item Compute the Maximum entropy estimator given by equation \eqref{mem}.
\end{itemize}
We then s¿display the resulting histogram in Figure 1.

\begin{center}
\begin{figure}[t!]
\epsfig{file=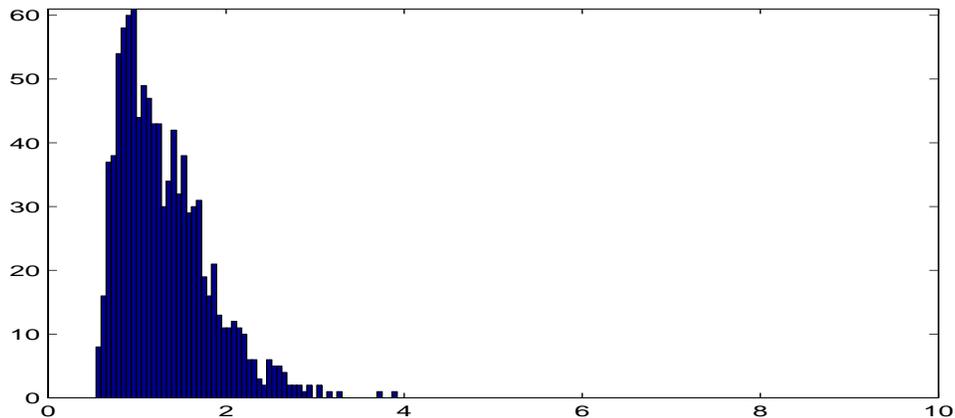,width=6in,height=2.5in}
\caption{Histogram for $E(x)$ with the Maximum Entropy Method.}
\label{fig:mem}
\end{figure}
\end{center}

\subsection{The bayesian estimator}
In this section we derive the algorithm for a Bayesian inference of the model given by $y_i = x + e_i$, for $i = 1,2,...,n$. The classical likelihood estimator of $x$ is given by $\hat{y}=\frac{1}{n}\sum_{i=1}^{n}y_i$. As we know that the unknown mean $x$ has an exponential probability distribution with parameter $\theta$ $(x\sim\mathbb{E}(\theta))$, therefore the joint density of the $y_i$ and $\mu$ is proportional to:
\begin{equation}
\prod_{i=1}^{n}\frac{1}{\sqrt{2\pi\delta^2}}\exp\left\{-\frac{(y_i - x)^2}{2\delta^2}\right\}\theta\exp (-\theta x)\pi (\theta)
\end{equation}
where $\theta\exp (-\theta x)$ is the density of the unknown mean $x$ and where $\pi (\theta)\propto \theta^{-1}$ is the Jeffrey's non informative prior distribution for the parameter $\theta$ Berger (1985).

In order to derive the Bayesian estimator, we need to get the posterior probability distribution for $\theta$, which we do with the following Gibbs sampling scheme, described in Robert and Casella (2005):
\begin{itemize}
\item Draw $x\sim N\left(\hat{y}-\frac{\theta\delta^2}{n},\frac{\delta^2}{n}\right)1_{x>0}$
\item Draw $\theta\sim\mathbb{E}(x)$
\end{itemize}
Repeat this algorithm many times in order to obtain a large sample from the posterior distribution of $\theta$ in order to obtain the posterior distribution of $E(x)=\frac{1}{\theta}$. For our application, we simulate data with $\theta = 1$, which gives an expected value for $x$ equal to $E(x)=1$.

We get the histogram displayed in Figure 2 for the estimations of $E(x)$ after $1000$ iterations when simulating data for $\theta = 1$.

\begin{center}
\begin{figure}[t!]
\epsfig{file=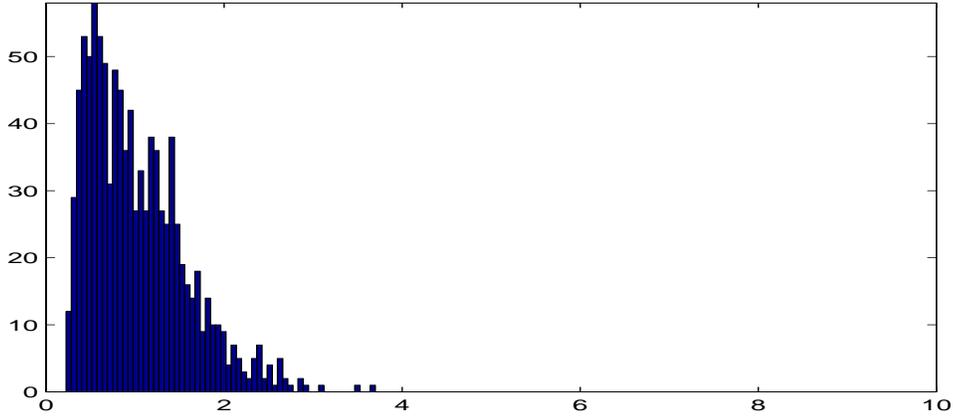,width=6in,height=2.5in} \caption{Histogram for $E(x)$ with Bayes Method.}
\label{fig:bayes}
\end{figure}
\end{center}
\subsection{The Maximum Likelihood estimator}
The problem of obtaining a ML estimator is complicated in this setup because data points are distributed like
\begin{eqnarray*}
f_\theta(t) &=& \int_{-\infty}^t \theta e^{-\theta(t-s)}e^{-s^2/2\delta^2}ds/\sqrt(2\pi\delta^2)\\
f_\theta(t) &=& \theta e^{-\theta t + \frac{(\theta\delta)^2}{2}}\mathbb{P}(S<t)
\end{eqnarray*}

where $S\sim N\left(\theta\delta^2 , \delta^2\right)$. Therefore, after observing $t_{1}$, $t_{2}$, and $t_{3}$, we get the following likelihood that we maximize numerically:
\begin{equation}\label{numerical}
\theta^3 e^{-\theta\sum_{i=1}^{3} t_{i} + \frac{3(\theta\delta)^2}{2}}\prod_{i=1}^{3}\mathbb{P}(S<t_{i}).
\end{equation}
If we attempted to obtain the ML estimator analytically, we would need to solve

$$\frac{n}{\theta}- \sum_{j=1}^n \frac{\int_{-\infty}^{t_j} \theta e^{-\theta(t_j-s)}e^{-s^2/2\delta^2}ds/\sqrt(2\pi\delta^2)}{\int_{-\infty}^{t_j} \theta e^{-\theta(t_j-s)}e^{-s^2/2\delta^2}ds/\sqrt(2\pi\delta^2)} = 0.$$

Notice that as $\delta \rightarrow 0$ this equation tends to $\frac{n}{\theta} - \sum_{j=1}^n t_j = 0$ as expected. We can move forward a bit, and integrate by parts each numerator, and after some calculations we arrive to

$$\frac{n}{\theta} - \sum_{j=1}^n t_j + n\delta^2\theta - \sum_{j=1}^n\frac{\delta e^{-t_j^2/2\delta^2}}{\int_{-\infty}^{t_j} \theta e^{-\theta(t_j-s)}e^{-s^2/2\delta^2}ds/\sqrt(2\pi\delta^2)} = 0.$$

Trying to solve this equation in $\theta$ is rather hopeless. That is the reason why we carried on a numerical maximization procedure on (\ref{numerical}). To understand what happens when the noise is small,  we drop the last term in the last equation and we are left with
$$\frac{n}{\theta} - \sum_{j=1}^n t_j + n\delta^2\theta$$
the solution of which is

$$\frac{1}{\theta}^* = \frac{1}{2}\big(\hat{y} + \sqrt{\hat{y}^2 - 4\delta^2}\big)$$

or $\theta^* = 2\big(\hat{y} + \sqrt{\hat{y}^2 - 4\delta^2}\big)^{-1},$ and we see that the effect of noise is to increase the ML estimator. In figure 3 we plot the histogram of $\frac{1}{\theta}^*$ obtained by numerically maximizing (\ref{numerical}) for each simulated data point.

\begin{center}
\begin{figure}[t!]
\epsfig{file=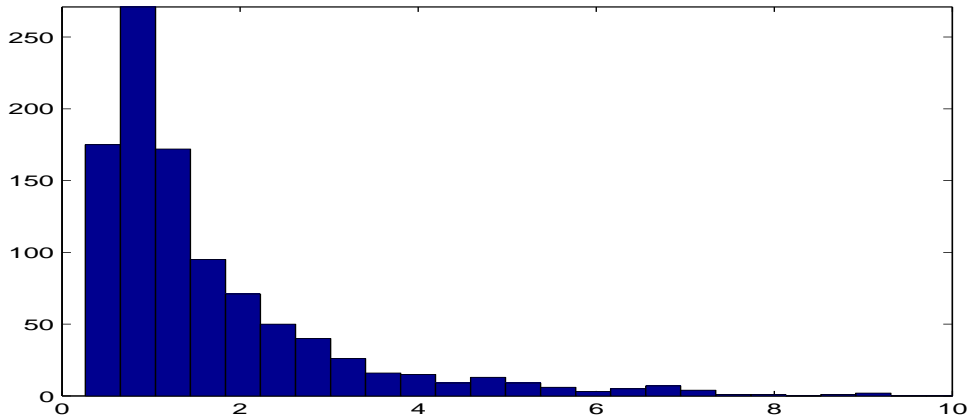,width=6in,height=2.5in} \caption{Histogram for $E(x)$ with the Maximum Likelihood Method.}
\label{fig:mle}
\end{figure}
\end{center}

When simulating data for $\theta = 1$, the MEM, Maximum likelihood and Bayesian histograms are all skewed to the right and yield a mean under the three simulated histograms close to 1. The MEM method yields a sample mean of $1.3252$ with a sample standard deviation of $0.5$, the Bayesian yields sample means equal to $1.045$ and sample standard deviation of $0.5529$, and the Maximum Likelihood method yields a sample mean of $1.81$ with a sample standard deviation of $2.29$. All the three methods produce right skewed histograms for $E(x)$. The MEM and Bayesian method provide better and similar results and more accurate than the Maximum Likelihood method.

\section{Concluding remarks}
On one hand, MEM backs up the intuitive belief, according to which, if the $y_i$ are all the data that you have, it is all right to compute your estimator of the mean for $\alpha =0$. The MEM and Bayesian methods yield closer results to the true parameter value than the maximum likelihood estimator.

On the other, and this depends on your choice of priors, MEM provides us with a way of modifying those priors, and obtain representations like $\hat{y} = \hat{x}^* + \hat{e}^*$; where of course $\hat{x}^* = \hat{x}^*(\hat{y})$. What we saw above, is that there is a choice of prior distributions such that $\hat{x}^* = \hat{y}$ and $\hat{e}* = 0.$

The important thing is that this is actually true regardless of what the ``true'' probability describing the $x_i$ is.


\end{document}